%% file: CVPR/arxiv.tex
\documentclass[10pt,twocolumn,letterpaper]{article}

\usepackage[pagenumbers]{cvpr} 

\input{preamble}
\definecolor{cvprblue}{rgb}{0.21,0.49,0.74}
\usepackage[pagebackref,breaklinks,colorlinks,allcolors=cvprblue]{hyperref}
\usepackage{multirow} 
\usepackage{caption}
\usepackage{cuted}
\usepackage{afterpage}

\def\paperID{14145} 
\def\confName{CVPR}
\def\confYear{2026}

\title{TEXTRIX: Latent Attribute Grid for Native Texture Generation and Beyond}

\author{
Yifei Zeng\textsuperscript{1,2*} \quad
Yajie Bao\textsuperscript{1,2*} \quad
Jiachen Qian\textsuperscript{3,2} \quad
Shuang Wu\textsuperscript{1,2} \quad
Youtian Lin\textsuperscript{1} \\
Hao Zhu\textsuperscript{1} \quad
Buyu Li\textsuperscript{4} \quad
Feihu Zhang\textsuperscript{2} \quad
Xun Cao\textsuperscript{1} \quad
Yao Yao\textsuperscript{1$\dagger$} \\[1.5ex] 
\textsuperscript{1}Nanjing University \quad
\textsuperscript{2}DreamTech \quad
\textsuperscript{3}HKU \quad
\textsuperscript{4}OriginArk \\
}
\begin{document}
\maketitle

{
    \renewcommand{\thefootnote}{}
    \footnotetext{\textsuperscript{*}Equal contribution \quad
    {$\dagger$}Corresponding author   }

}

\begin{strip}
	\vspace{-40pt}
	\centering
	\includegraphics[width=1\textwidth]{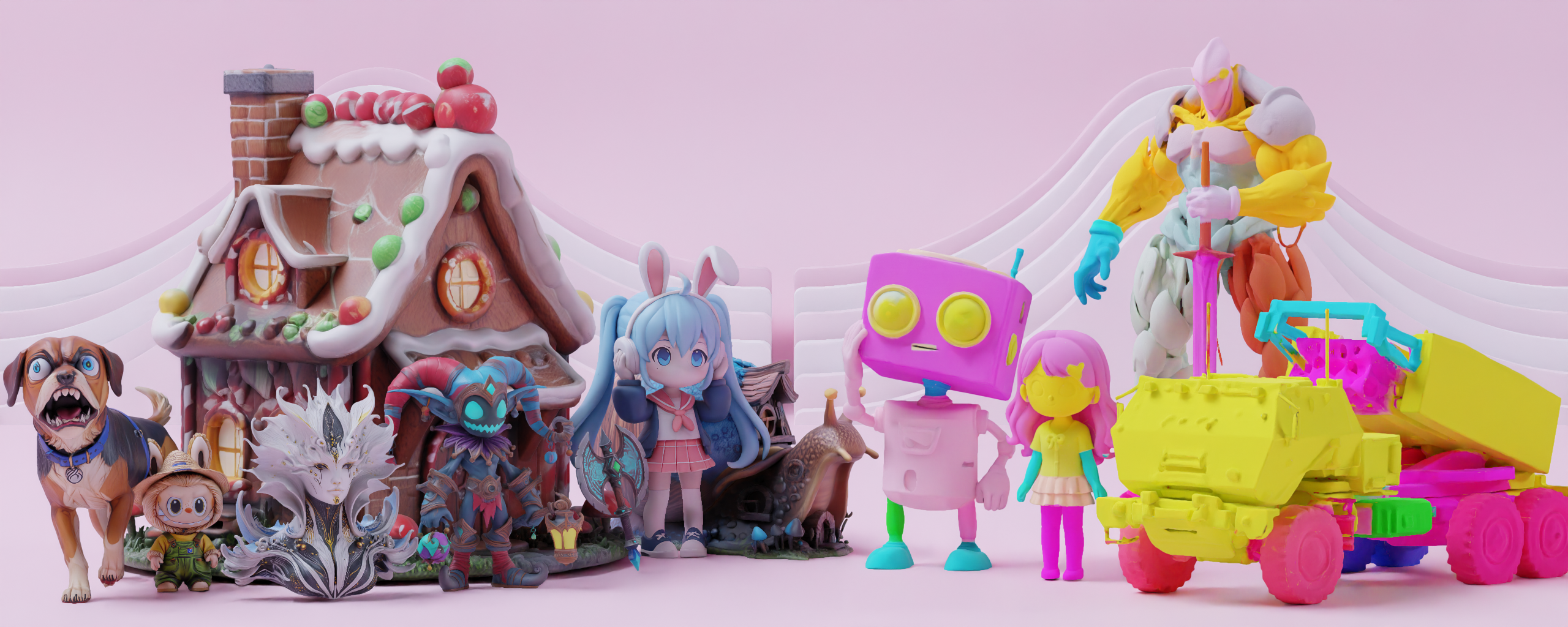}
    \vspace{-8pt}
    \captionsetup{type=figure,font=small,position=top}
    \caption{
        Seamless texturing and precise segmentation with our TEXTRIX framework. Our model generates geometrically aligned textures and segmentations from a single-view input, avoiding the inter-view inconsistencies that commonly affect prevailing 3D generation.
    } 
    \label{fig:teaser}
    \vspace{-20pt}
\end{strip}

\begin{abstract}
Prevailing 3D texture generation methods, which often rely on multi-view fusion, are frequently hindered by inter-view inconsistencies and incomplete coverage of complex surfaces, limiting the fidelity and completeness of the generated content. To overcome these challenges, we introduce TEXTRIX, a native 3D attribute generation framework for high-fidelity texture synthesis and downstream applications such as precise 3D part segmentation. Our approach constructs a latent 3D attribute grid and leverages a Diffusion Transformer equipped with sparse attention, enabling direct coloring of 3D models in volumetric space and fundamentally avoiding the limitations of multi-view fusion. Built upon this native representation, the framework naturally extends to high-precision 3D segmentation by training the same architecture to predict semantic attributes on the grid. Extensive experiments demonstrate state-of-the-art performance on both tasks, producing seamless, high-fidelity textures and accurate 3D part segmentation with precise boundaries. Project Page: \url{www.neural4d.com/research-page/textrix}
\end{abstract}
\enlargethispage{2\baselineskip}

\input{CVPR/section2/intro}
\input{CVPR/section2/related_works}
\input{CVPR/section2/method}
\input{CVPR/section2/implement}
\input{CVPR/section2/exp}

\input{CVPR/section2/limit_and_conclusion}
{
    \small
    \bibliographystyle{ieeenat_fullname}
    \bibliography{CVPR/main}
}


\end{document}

%% file: CVPR/section2/intro.tex
\section{Introduction}

High-fidelity 3D assets are foundational to immersive virtual environments, digital media, and robotic simulation. While 3D geometry generation has seen significant progress, applying realistic and coherent textures remains a critical bottleneck to 3D generation. The quality of 3D texturing is often compromised by the underlying 2D-centric paradigms used to create them.

A dominant line of 3D generation attempts to lift powerful 2D diffusion models for synthesis. These methods either optimize a 3D representation through score distillation~\cite{poole2022dreamfusion,lin2022magic3d,chen2023fantasia3d,wang2023prolificdreamer,tang2023dreamgaussian,tang2023makeit3d,Magic123} or generate multi-view images and reconstruct 3D from them~\cite{liu2023zero1to3,tang2023MVDiffusion,shi2023MVDream,wang2023imagedream,instant3d2023,lu2024direct2,tang2024lgm,xu2024grm}. Recent variants fuse these multi-view projections onto a mesh~\cite{huang2024mvadapter,hunyuan3d22025tencent}, yet the paradigm remains fundamentally limited: inter-view inconsistencies introduce seams and blurred details, lighting becomes incoherent, and complex or occluded regions frequently lack plausible coverage.

\begin{figure}[htbp]
  \centering

  \includegraphics[width=\columnwidth]{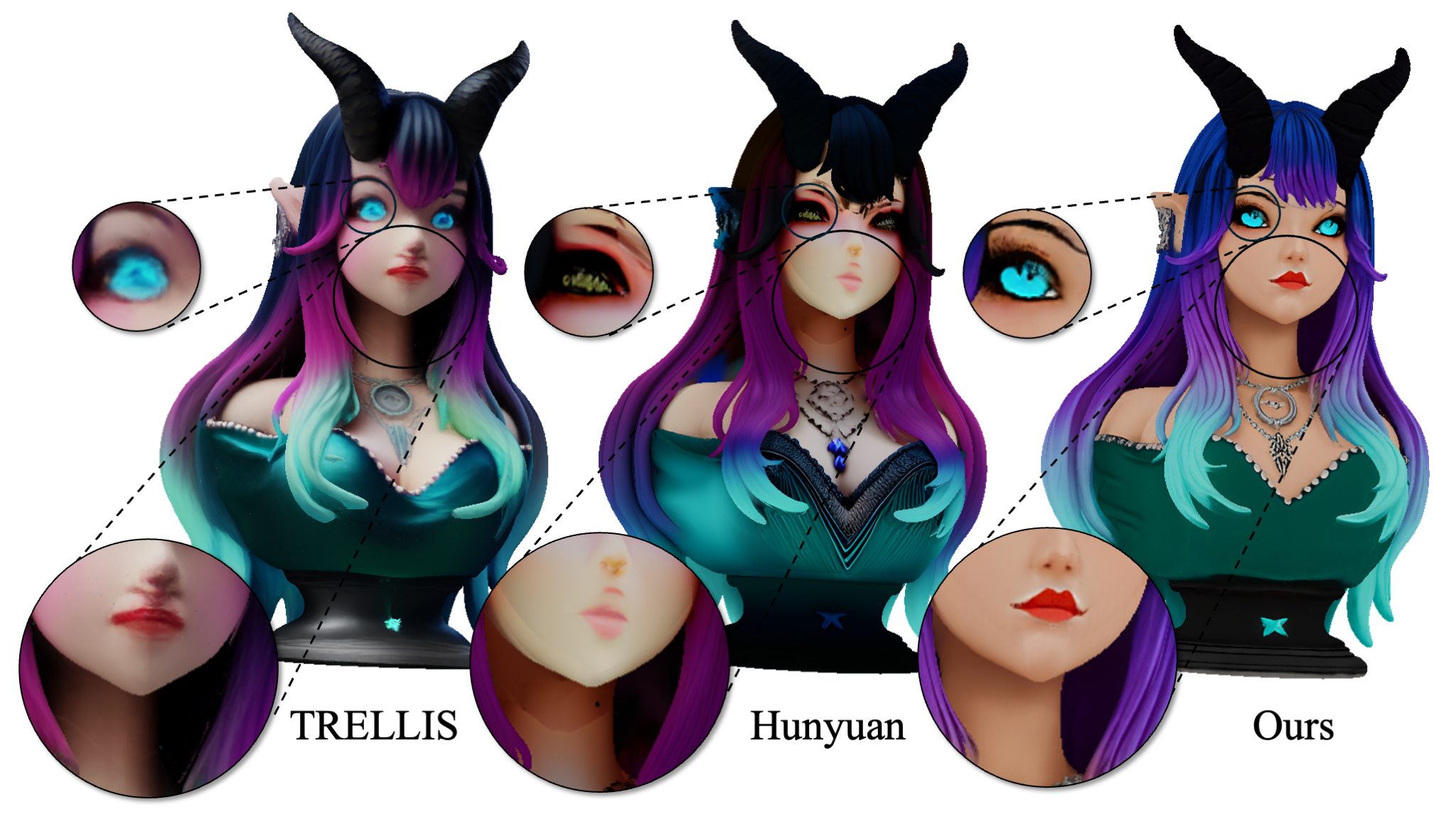}
  \caption{Limitations of prevailing paradigms. \textbf{Left}: TRELLIS exhibits inconsistency and low fidelity on the lips. \textbf{Middle}: Multi-view fusion of Hunyuan3D 3.0 results in noticeable seams and blurring around the mouth and neck. \textbf{Right}: Our native 3D framework (TEXTRIX) produces a seamless, high-fidelity result.}

  \label{fig:vae}
\end{figure}

To circumvent these fusion issues, other methods explore generation directly in the 2D UV parameterization space \cite{richardson2023texture, chen2023text2tex, zeng2024paint3d,texgen}. While this avoids per-object optimization, we posit that this paradigm is still not truly \textit{native} to the 3D domain. The UV mapping process itself flattens a continuous 3D surface into a set of disjointed 2D islands, severing natural geodesic neighborhoods. Models trained on this fragmented representation struggle to enforce 3D global coherence, resulting in artifacts at UV seams and a disconnect between the learned texture patterns and the 3D geometry they are meant to describe.

In this work, we propose a new paradigm that addresses these limitations by operating directly in 3D space. Our core idea is to represent and generate texture as a native 3D attribute grid—a sparse voxel field that stores diverse information. We leverage a Diffusion Transformer (DiT) \cite{Peebles2022DiT}, empowered by sparse attention, to learn the distribution of these 3D grids. This native 3D approach, which builds on recent advances in generating 3D shapes with sparse \cite{ren2024xcube} or structured \cite{xiang2024structured} voxel representations, fundamentally bypasses the problems of both multi-view inconsistency and 2D UV fragmentation, enabling the synthesis of seamless, high-resolution, and geometrically-aware textures.

A key challenge in generative texturing is ensuring the output faithfully aligns with a user-provided condition. To achieve robust frontal alignment, we introduce a novel sparse latent conditioning strategy. Instead of relying solely on global features (e.g., CLIP embeddings \cite{Radford2021LearningTV}), we project the input image into a 3D voxel grid, creating a sparse latent condition. This condition, which is encoded and fed into the DiT via fused cross-attention, provides a strong, spatially-explicit prior that robustly anchors the generative process. This ensures the generated texture not only matches the input view's details with high fidelity but also uses that information to plausibly complete occluded regions.

A significant advantage of our framework is its inherent flexibility. The 3D attribute grid is a unified representation not limited to color. By training the same architecture to predict semantic information, our framework naturally extends to tasks like high-precision 3D segmentation and PBR generation. We demonstrate that our model achieves state-of-the-art (SOTA) performance on downstream 3D part segmentation tasks, validating the power and potential of our native 3D paradigm for unifying various tasks. Our contributions are threefold:
\begin{itemize}
\item A native 3D texturing framework, built upon a sparse attribute grid, that generates high quality textures by resolving the inter-view inconsistencies and seam artifacts from traditional methods.
\item A novel sparse latent conditioning strategy that projects visual information into the 3D grid, achieving high-fidelity frontal alignment and detail preservation.
\item An extensible framework for 3D segmentation generation and beyond, achieving SOTA results by training the same sparse grid architecture to predict diverse attributes. 
\end{itemize}

\afterpage{
  \clearpage 
  \begin{figure*}[h!] 
    \centering 
    \includegraphics[height=0.98\textheight]{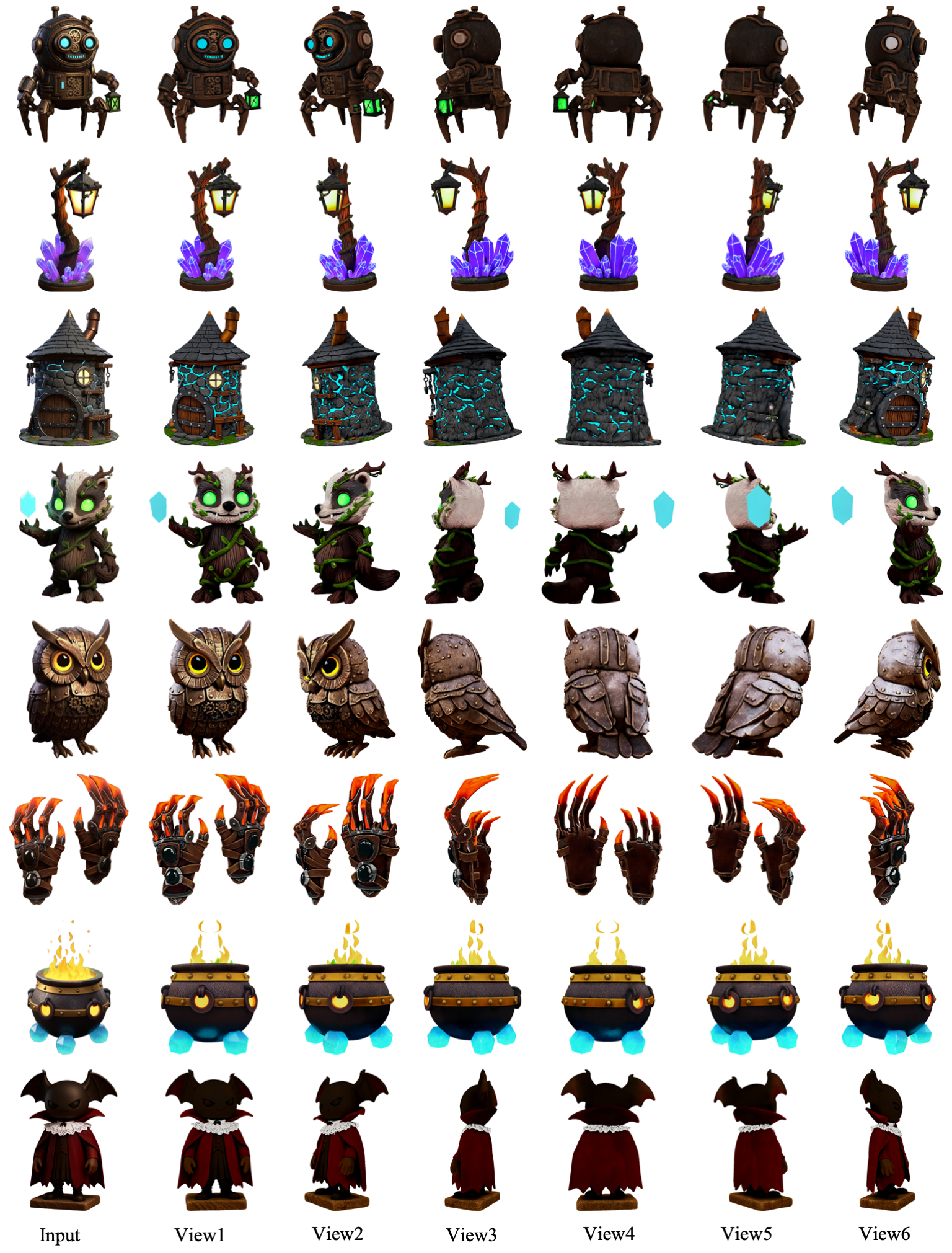}
    \caption{Visualization for Our Seamless Single-View Generation.}
    \label{fig:qual_texture}
  \end{figure*}
  \clearpage 
}

%% file: CVPR/section2/related_works.tex
\section{Related Works} 
Our work is positioned at the intersection of 3D generative modeling, texture generation and 3D segmentation. We build upon recent advances in these fields, while proposing a general framework that addresses the limitations of existing specialized approaches. 

\subsection{Native 3D Generation} 

Native 3D generative models learn directly from large-scale 3D datasets\cite{objaverse,objaverseXL,yu2023mvimgnet}, leading to higher-fidelity geometric outputs.
Recent efforts have shifted from early works on GANs~\cite{Chan2021} towards diffusion models operating on diverse 3D representations such as point clouds~\cite{luo2021diffusion}, implicit fields~\cite{cheng2023sdfusion}. To enhance quality and efficiency, the field has converged on learning within a compact latent space. These latent diffusion models 
can be broadly categorized as implicit or explicit. Implicit methods, such as those based on vecsets or SDFs~\cite{zhang2024clay,3DShape2VecSet,direct3d,Chen_2025_Dora,li2025triposg}, excel at representing continuous surfaces but can be complex to train and decode. Explicit methods, which utilize representations like sparse voxels, offer more direct control and interpretability~\cite{ren2024xcube,xiang2024structured}. However, they face significant computational and memory challenges limiting their output resolution, though recent works like Direct3D-S2~\cite{wu2025direct3ds2gigascale3dgeneration} begin to address this scalability with efficient attention mechanisms for sparse data. Our work contributes to this native 3D paradigm by introducing a volumetric color grid that serves as a versatile and direct representation for both generation and perception.

\subsection{Texture Generation} 
Applying realistic appearance to 3D geometry is a critical task, which has been predominantly addressed by leveraging powerful pre-trained 2D text-to-image (T2I) models. One major line of work involves test-time optimization, where multi-view images are generated and projected onto the mesh surface. This line of work includes methods for progressive inpainting~\cite{richardson2023texture,chen2023text2tex}, synchronized multi-view diffusion~\cite{cao2023texfusion}, and adapter-based solutions to improve consistency~\cite{huang2024mvadapter}. Despite these advances, these projection-based techniques are prone to artifacts, slow per-instance optimization, and incomplete surface coverage. 

Another approach attempts to circumvent projection issues by generating textures directly in a 2D UV parameterization space, for instance by training a diffusion model to produce UV maps in a feed-forward manner~\cite{texgen}. While this avoids multi-view fusion, it introduces problems inherent to UV mapping: the 2D representation discards 3D geodesic information, leading to seams and discontinuities at the boundaries of UV islands. Our framework addresses the core limitations of both approaches. By generating a native 3D color field and querying it with 3D coordinates, we completely bypass the pitfalls of both multi-view projection and UV space fragmentation, enabling the synthesis of seamless, high-quality textures. 

\subsection{3D Part Segmentation} 
3D part segmentation is a fundamental perception task, but supervised methods~\cite{qi2017pointnetplusplus} are limited by data scarcity. Recent work distills 2D foundation models, often using text prompts (VLMs)~\cite{liu2022partslip}. However, text can be ambiguous and fails to describe many geometric or functional parts.

A more general, class-agnostic direction lifts 2D SAM masks~\cite{kirillov2023segany}, but state-of-the-art methods now train feed-forward networks (e.g., SAMPart3D~\cite{yang2024sampart3d}, PartField~\cite{partfield2025}) to learn continuous part-aware feature fields. Our work aligns with this native 3D approach. By operating on a high-resolution volumetric grid, our framework preserves fine-grained geometric details, producing sharper and more accurate segmentation boundaries.

%% file: CVPR/section2/method.tex
\section{Method}

\begin{figure*}[!t]
  \centering
  \includegraphics[width=0.95\linewidth]{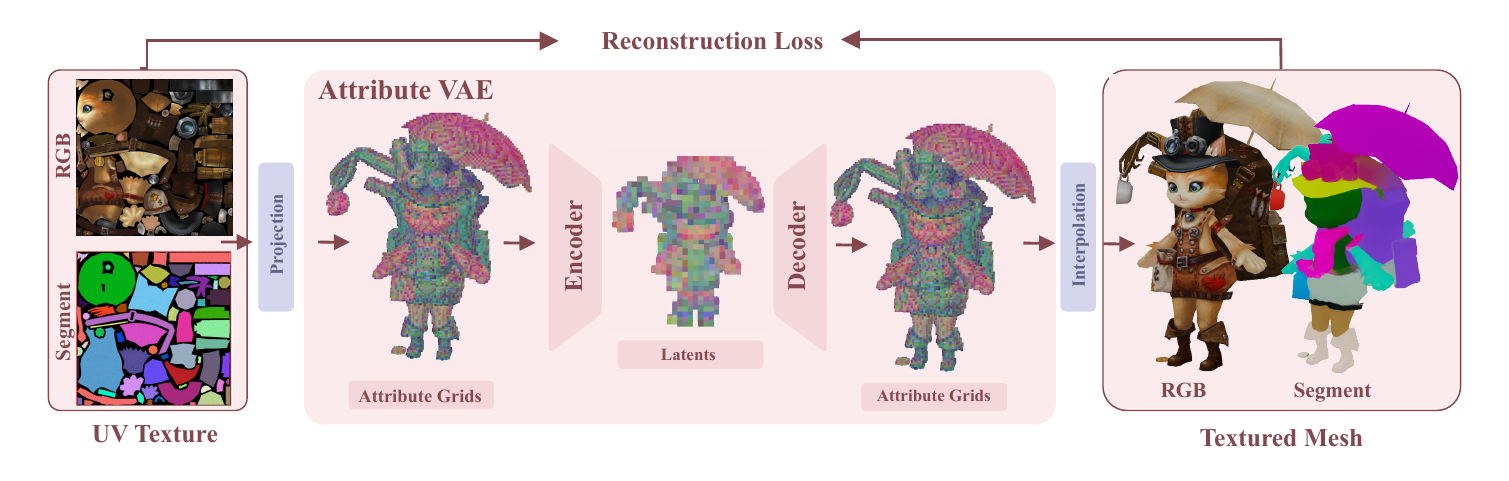}
  \caption{The VAE of our method. We introduce a native latent 3D attribute representation, which contains properties such as color, semantic labels, and PBR materials of the object within each sparse voxel. An end-to-end attribute VAE is employed to encode the representation into a continuous and compact latent space.}
  \label{fig:pipeline_vae}
\end{figure*}



Traditional approaches to 3D generation and perception tasks have typically treated these problems separately, with distinct frameworks designed for each. In this work, we propose a unified and native 3D architecture capable of supporting high-resolution 3D content generation as well as high-precision 3D perception. Central to our approach is the introduction of a native 3D attribute representation, which stores properties such as color, semantic labels, and PBR materials within a sparse voxel field (Sec~\ref{sec:representation}). We then design an efficient attribute variational autoencoder (VAE) to learn a compact latent space, ensuring high-quality reconstruction (Sec~\ref{sec:vae}). Building on this latent space, we devise an image-conditioned attribute diffusion transformer (DiT) for generation and perception tasks, with a sparse latent conditioning strategy to enhance alignment between the generated content and the input images (Sec~\ref{sec:dit}). Finally, we demonstrate the versatility of our unified representation by extending the framework to high-precision tasks, including 3D part segmentation and PBR material generation (Sec~\ref{sec:seg}).

\subsection{Latent 3D Attribute Representation}
\label{sec:representation}

Representing 3D attributes is a fundamental challenge for both 3D generation and perception tasks. Prior approaches, such as TRELLIS~\cite{xiang2024structured}, rely on pre-trained DINOv2~\cite{oquab2023dinov2} encoder to extract features from multi-view images, which are subsequently projected into corresponding voxels as appearance representations. However, this approach is fraught with several limitations. Firstly, the use of DINOv2 features results in a significant loss of high-frequency information, which diminishes accuracy. Furthermore, the high dimensionality of the feature channels introduces substantial computational overhead, making it difficult to scale to higher resolutions, such as $1024^3$. In addition, this representation is inherently limited in its ability to extend to other 3D attributes, such as PBR materials.

\noindent\textbf{Native 3D Attribute Grid.} To address these issues, we introduce a novel native 3D attribute representation. Using a sparse voxel grid structure, we store high-resolution raw 3D attributes in regions adjacent to the object’s surface, thereby improving both efficiency and scalability. To be specific, the native 3D attributes $\mathcal{A}$ of a given object is represented as:
\begin{align}
    &\label{eq:attribute}
    \mathcal{A} = \left\{ \mathbf{a}_i \,\middle|\, \mathbf{a}_i \in \mathbb{R}^{k} \right\}_{i=1}^{M}, \\
&k = k_{\text{color}} + k_{\text{semantic}} + k_{\text{PBR}} + \cdots + k_{\text{e}},
\end{align}
where $\mathbf{a}_i$ denotes the attributes vector stored at each voxel, $M$ is the total number of the sparse voxels, $k$ is the total channel dimension of the attribute space, $k_{\text{color}}$, $k_{\text{semantic}}$, $k_{\text{PBR}}$ and $k_{\text{e}}$ denote the channel dimension of color, semantic, PBR materials and other 3D attributes, respectively. 
\noindent\textbf{Querying via UV Position Map.} To query the attributes at any point on the model's surface, we first establish a mapping from the 2D UV attribute space to the 3D object space. This is accomplished by pre-calculating a UV position map, a high-resolution 2D image in which each pixel $(u, v)$ stores the corresponding 3D world coordinate $\mathbf{p} = (x, y, z)$ on the mesh surface. This map effectively functions as a lookup table, enabling the precise identification of the 3D point $\mathbf{p}$ for any given 2D attribute coordinate.
Once the 3D query point $\mathbf{p} \in [-0.5, 0.5]^3$ is obtained, we retrieve its attributes from the sparse voxel grid using standard trilinear interpolation. The procedure is outlined as follows: Initially, the coordinates of the query point are scaled to align with our grid of resolution $R$, thereby identifying the specific voxel cell that contains it. The base index of this cell's origin corner is given by $\mathbf{v}_0 = \lfloor \mathbf{p} \cdot R \rfloor$. Subsequently, we identify the eight corners of the enclosing voxel, denoted as:
\begin{equation}
    V = \{\mathbf{v}_{ijk} | i,j,k \in \{0,1\}\},
\end{equation}
and retrieve the corresponding attributes $\mathbf{a}_{ijk} \in \mathbb{R}^3$ for each corner from the sparse grid. Finally, the attributes at the query point $\mathbf{A}(\mathbf{p})$ are computed by performing trilinear interpolation on the eight corner attributes, weighted according to the query point’s local coordinates within the voxel:
\begin{align}
        &\mathbf{A}(\mathbf{p}) = \sum_{i,j,k \in \{0,1\}} (1-\boldsymbol{\alpha}_i)(1-\boldsymbol{\alpha}_j)(1-\boldsymbol{\alpha}_k) \mathbf{a}_{ijk}, \\
        &\boldsymbol{\alpha} = (\mathbf{p} \cdot R) - \mathbf{v}_0.
\end{align}
This simple yet effective approach enables efficient and continuous attribute querying from a compact, sparse representation, thereby forming the core mechanism of our native 3D framework.

This adaptive interpolation scheme offers a distinct advantage over conventional methods by endowing the model with the flexibility to learn a non-linear attribute field, thereby yielding substantially higher generation fidelity and perception precision even from a sparse representation.

\subsection{Attribute VAE}
\label{sec:vae}

\noindent\textbf{End-to-End Framework.} Methods such as TRELLIS~\cite{xiang2024structured}, which rely on non-native representations, use DINOv2 features extracted from multi-view images as input to their VAE, and require decoding into FlexiCubes~\cite{shen2023flexible} or 3D Gaussian splatting~\cite{kerbl20233dgs} representations to supervise appearance. This approach suffers from inefficiencies in training and is inherently limited by a resolution of $256^3$. In contrast, leveraging our flexible and efficient native 3D attribute representation, we propose a fully end-to-end, symmetric attribute VAE, which is shown in Figure~\ref{fig:pipeline_vae}. This architecture not only circumvents the limitations of prior methods but is also capable of scaling to higher resolutions, such as $1024^3$.
\begin{figure*}[!t]
  \centering
  \includegraphics[width=0.95\linewidth]{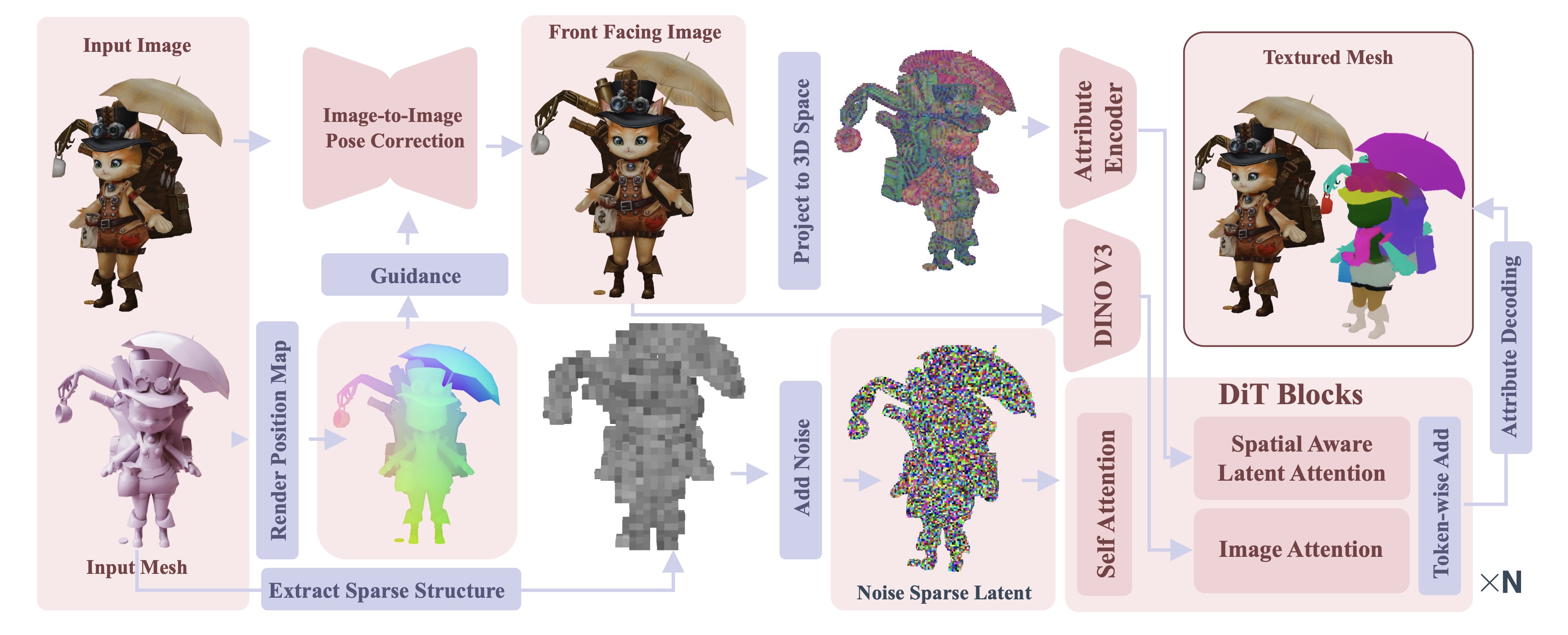}
  \caption{Architecture of our Diffusion Transformer (DiT). This image-conditioned model operates on sparse latents to perform unified generation (texturing) and perception (segmentation). We introduce a novel sparse latent conditioning strategy that ensures high-fidelity alignment with the input image.}
  \label{fig:pipeline_dit}
\end{figure*}

\label{sec:dit}
\noindent\textbf{Network Architecture.} The VAE adopts a hybrid architecture that integrates the complementary strengths of sparse 3D convolutions and transformers. The encoder begins by processing the input sparse attribute grid through a hierarchy of sparse convolutional modules, which progressively downsample the spatial resolution while extracting multi-scale local features. The resulting coarse feature grid is then serialized into a sequence of tokens and passed through a sparse transformer backbone, enabling the modeling of long-range dependencies and global contextual relationships. This stage effectively maps the high-dimensional input into a compact latent distribution. Symmetrically, the decoder mirrors this process: a sparse transformer first interprets a sampled latent code, after which a series of sparse transposed convolutional modules progressively upsample the spatial representation, ultimately reconstructing a detailed, high-resolution attribute grid.

\noindent\textbf{Training Objective.} The decoded sparse attribute grid contains both valid and redundant regions. To mitigate this, we introduce a pruning operation applied after each upsampling stage. Specifically, the input grid is sequentially downsampled by factors of 2, 4, and 8, after which grid elements are selectively removed based on their spatial correspondence to valid surface regions. To further enhance the fidelity of the reconstructed attribute grid, we avoid direct supervision from the input grid and instead adopt an online rendering paradigm, where view-dependent renderings provide attribute supervision. Additionally, a Kullback–Leibler (KL) divergence loss is imposed to regularize the latent space, while perceptual (LPIPS) and adversarial (GAN) losses are incorporated to further refine visual realism and reconstruction quality. The final loss of the sparse VAE is a weighted combination:
\begin{equation}
\label{eq:vae_loss}
    \mathcal{L}_{\text{total}} = \lambda_{1} \mathcal{L}_{1} + \lambda_{\text{prune}}\mathbf{L}_{\text{prune}} + \lambda_{\text{kl}} \mathcal{L}_{\text{kl}} + \lambda_{\text{lpips}} \mathcal{L}_{\text{lpips}} + \lambda_{\text{adv}} \mathcal{L}_{\text{adv}},
\end{equation}
where $\mathcal{L}_{1}$ denotes the L1 loss, and $\mathcal{L}_{\text{prune}}$ is a binary cross-entropy (BCE) loss for pruning.

\subsection{Attribute DiT with Sparse Latent Conditioning}
Leveraging the latent space encoded by the attribute VAE, we train an image-conditioned diffusion transformer (DiT) to generate 3D attributes. The framework of our DiT is shown in Figure~\ref{fig:pipeline_dit}. Previous approaches typically employ pretrained visual encoders, such as CLIP or DINO, to extract image features as conditions. However, due to the inherent domain gap between these feature embeddings and the latent representation, the resulting attributes often exhibit misalignment with the input images. 

To address this issue, we propose a novel sparse latent conditioning mechanism. Specifically, to handle reference images $I$ from arbitrary, unknown viewpoints, we first canonicalize the input. We employ a finetuned diffusion model, conditioned on a position map rendered from the target front view, to synthesize a new front-view image $I_{\text{front}}$ based on the reference $I'$. Once this aligned $I_{\text{front}}$ is generated, we project its pixels onto a 3D volume using the same front-view position map. This projection creates an attribute grid that serves as the input condition. This sparse voxel grid is then fed into the encoder of our pretrained 3D attribute VAE, embedding the condition within the same latent space as the latent tokens $\mathbf{z}$, thus bridging the domain gap between the condition and the latent space. 

In addition to the sparse latent encoding, we also incorporate global semantic features extracted from a pretrained DINOv3 model~\cite{simeoni2025dinov3} as an additional condition to enhance model stability. This results in a hybrid cross-attention mechanism that injects both types of information into the diffusion process. The cross-attention is formulated as follows:
\begin{equation}
\begin{split}
    \hat{\mathbf{z}} = \text{CrossAttn}(\mathbf{z},\text{PE}(E_\text{VAE}(\mathcal{A}(I)))) \\ +  \text{CrossAttn}(\mathbf{z}, E_\text{DINOv3}(I)),
\end{split}
\end{equation}
where $\text{PE}$ represents position embedding, $E_\text{VAE}$ is the encoder of our attribute VAE, $\mathcal{A}(I)$ is the projected sparse volume of input images, and $E_\text{DINOv3}$ denotes the encoder of DINOv3.

Additionally, we adopt a configuration similar to Direct3D-S2~\cite{wu2025direct3ds2gigascale3dgeneration}, leveraging a spatial sparse attention mechanism to enhance the computational efficiency of self-attention. And we employ the rectified flow objective~\cite{esser2024scaling, lipman2022flow} to train the DiT.

\subsection{Applications and Extensibility}
\label{sec:seg}
We extend our framework's capabilities from generation to different challenge tasks, proposing that our unified representation can bridge these traditionally distinct domains. The native 3D attribute grid is not limited to color; it is a flexible structure designed to store diverse properties such as semantic labels and PBR materials. This inherent flexibility allows the same architecture to be adapted for various high-precision generation and perception tasks.

As a primary application, we adapt our model for 3D part segmentation. We frame this as a generative task: the model is trained to predict a latent attribute grid representing part labels rather than colors. During training, we use ground-truth UV maps where each distinct part is assigned a unique, random RGB color as a discrete label.

At inference, to generate the conditional input, we first utilize an existing 2D segmentation method \cite{chen2024subobject} to generate an initial, often oversegmented set of part masks from rendered views. To consolidate these fragmented regions into semantically meaningful components, we introduce a merging step based on feature similarity. We extract features for each adjacent segment pair using a DINO model and use the features to merge the similar pairs. Conditioned on this input, the model generates the final color-coded attribute grid. By querying this grid, we produce a UV map that directly corresponds to the 3D segmentation. Next, clustering methods are employed over label values to obtain appropriate segmentation results. A significant advantage of this native 3D approach is its ability to produce exceptionally sharp and precise segmentation boundaries, even on topologically complex meshes with a very high face count. 

Furthermore, our framework naturally extends to other generative tasks. For example, by training the model to predict attributes such as roughness, metallic, or normal maps within the same sparse grid structure, it can be seamlessly applied to PBR material generation, demonstrating the versatility of our approach. Results for PBR generation are presented in the Figure ~\ref{fig:qual_pbr}.

%% file: CVPR/section2/implement.tex
\section{Implementation Details}

\begin{figure*}[htbp]
  \centering
  \includegraphics[width=0.95\textwidth]{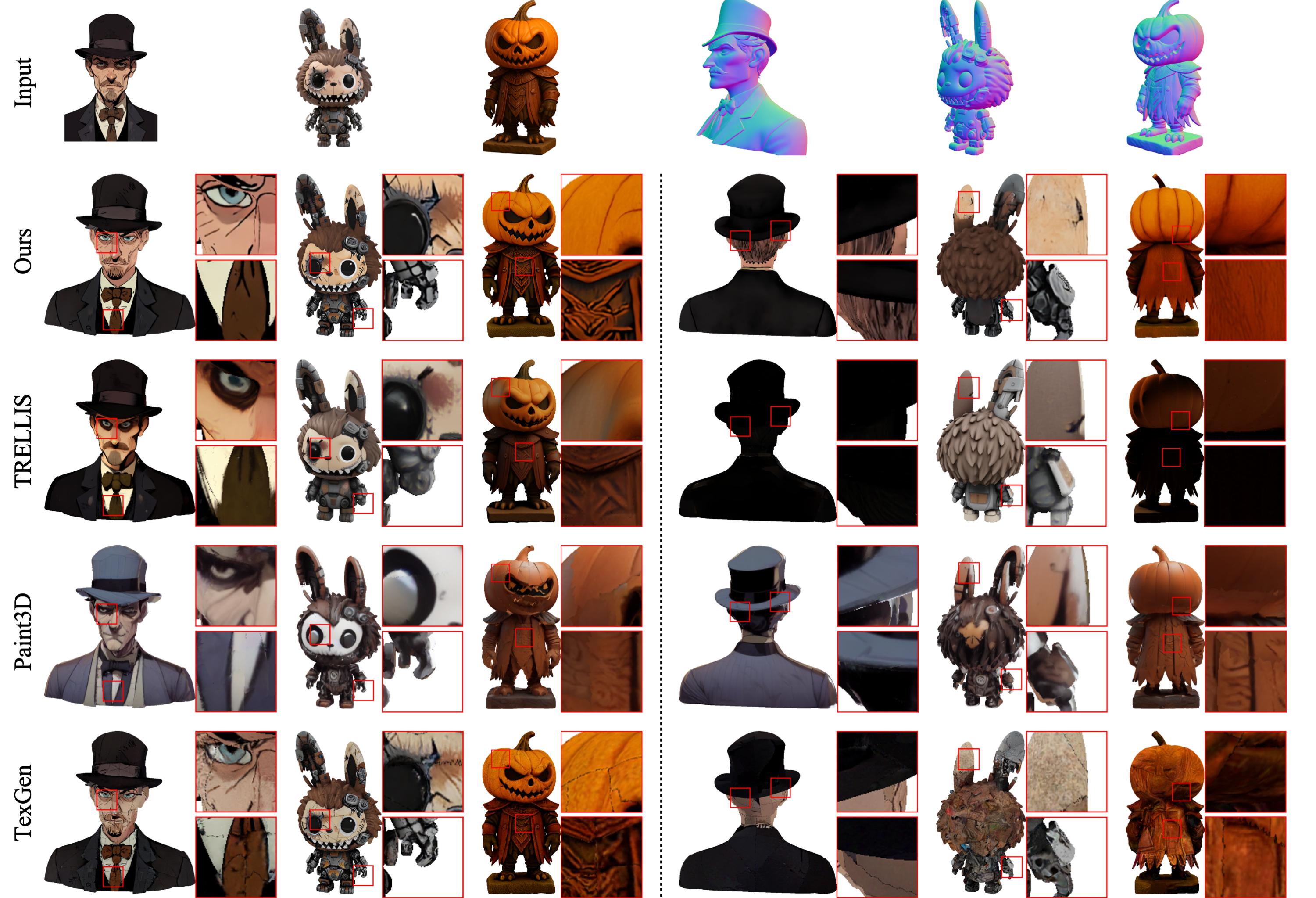}
  
  \caption{Qualitative comparison of single-view texture generation. Our method demonstrates superior fidelity, consistency with the input view, and freedom from artifacts compared to existing SOTA methods.}
  \vspace{-10pt}
  \label{fig:qual_texture}
\end{figure*}

\noindent\textbf{Dataset.} Our model is trained on a curated dataset derived from Objarverse\cite{objaverse}, ObjarverseXL\cite{objaverseXL} and private data. For each asset, we generate color-coded UV maps corresponding to texture and segmentation at a resolution of $2048 \times 2048$ to serve as training targets. We also render corresponding conditional images at the same resolution.

\noindent \textbf{Sparse VAE.} We train a Sparse VAE to encode both the texture map and the segmentation map into a compact latent space. The VAE compresses the input sparse grid from a resolution of $1024^3$ down to a latent grid of $128^3$, with a final latent embedding dimension of 16.

\noindent \textbf{Latent Diffusion Model.} Our core generative model is a sparse transformer-based architecture operating in the latent space. For conditioning, we use the pre-trained DinoV3\cite{simeoni2025dinov3} model to extract global image features. During training, we apply classifier-free guidance\cite{Ho2022ClassifierFreeDG} with a drop probability of 0.1 for the conditional inputs.

\noindent \textbf{Training.} We train the model end-to-end using the AdamW optimizer with a learning rate of $1 \times 10^{-4}$ and betas of $(0.9, 0.999)$. The training is performed with a batch size of 16 per GPU on 32 NVIDIA A100 GPUs. We use a flow matching scheduler with 1000 training timesteps. The entire model is trained for approximately 30,000 iterations. For inference, we use the same scheduler with 15 steps and a guidance scale of 3.0.

%% file: CVPR/section2/exp.tex
\section{Experiments}

%

To validate the efficacy of our framework, we conduct a comprehensive set of experiments targeting both 3D content generation and perception. We perform rigorous qualitative and quantitative comparisons against state-of-the-art methods in texture generation and part segmentation. Furthermore, we present detailed ablation studies to analyze the contribution of our key architectural components.

\subsection{Qualitative Results}

\paragraph{Single-View Texture Generation.} We first evaluate the fidelity of texture generation from a single input view. As illustrated in Figure \ref{fig:qual_texture}, we compare our method against leading approaches. Paint3D \cite{zeng2024paint3d} struggles with low-fidelity results, pronounced seams, and poor consistency with the input image. TexGen \cite{texgen} generates reasonable results for visible surfaces but suffers from chaotic and incoherent textures on occluded regions. Similarly, TRELLIS \cite{xiang2024structured} is prone to discoloration artifacts and fails to reproduce fine-grained frontal details. In contrast, our method demonstrates superior performance, generating high-quality, high-resolution textures that are tightly aligned with the input view. Our attribute grid effectively avoids the seams and inconsistencies that plague multi-view fusion and 2D UV-space methods.

\begin{figure}[htbp]
  \centering
  \includegraphics[width=\columnwidth]{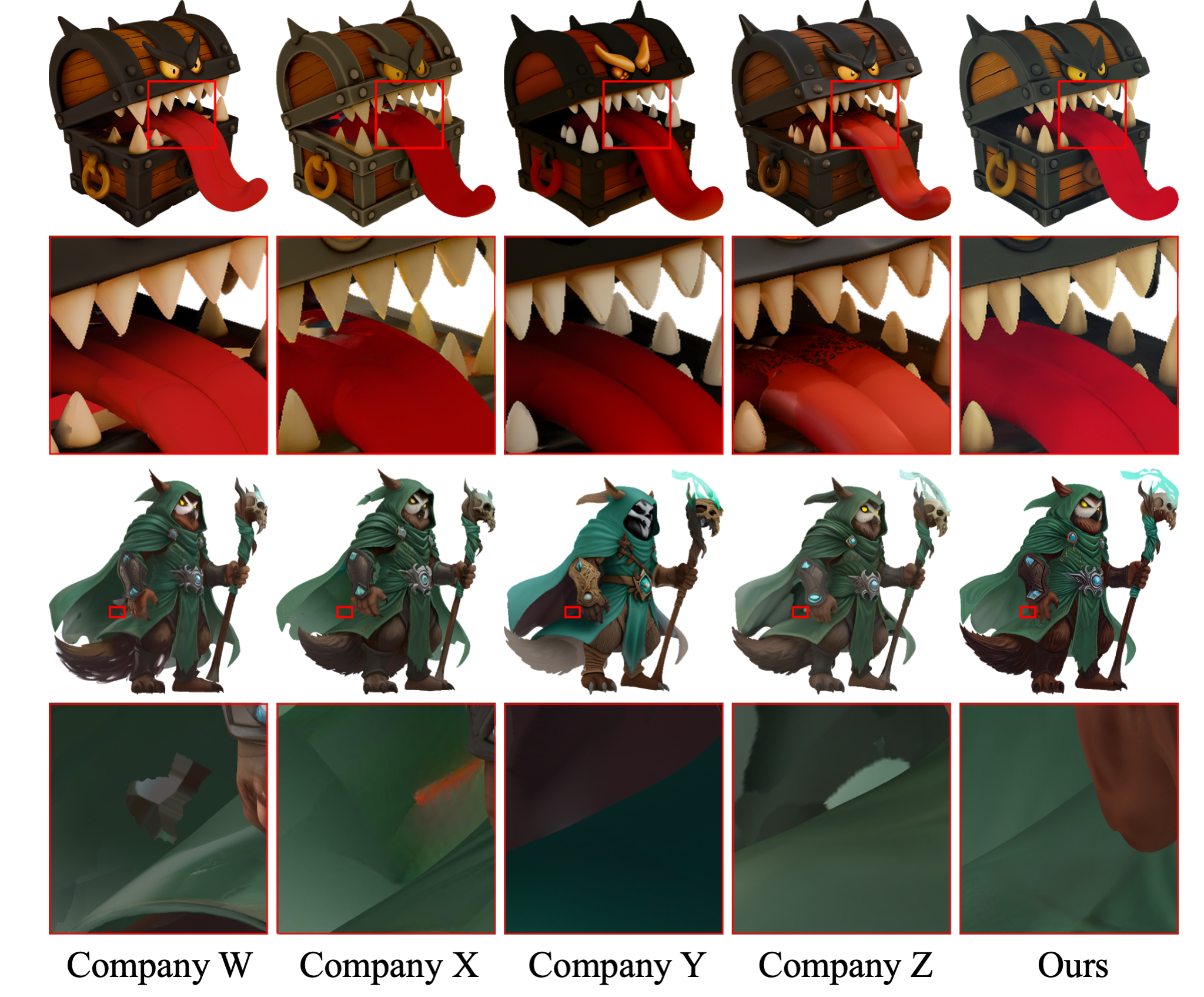}
  
  \caption{Qualitative comparison of multi-view conditioned generation compared to commercial methods. Our method demonstrates seamless results with fewer artifacts in the occluded region.}
  \label{fig:commercial}
\end{figure}
\vspace{-4mm}

\begin{figure*}[htbp]
  \centering
  \includegraphics[width=0.9\textwidth]{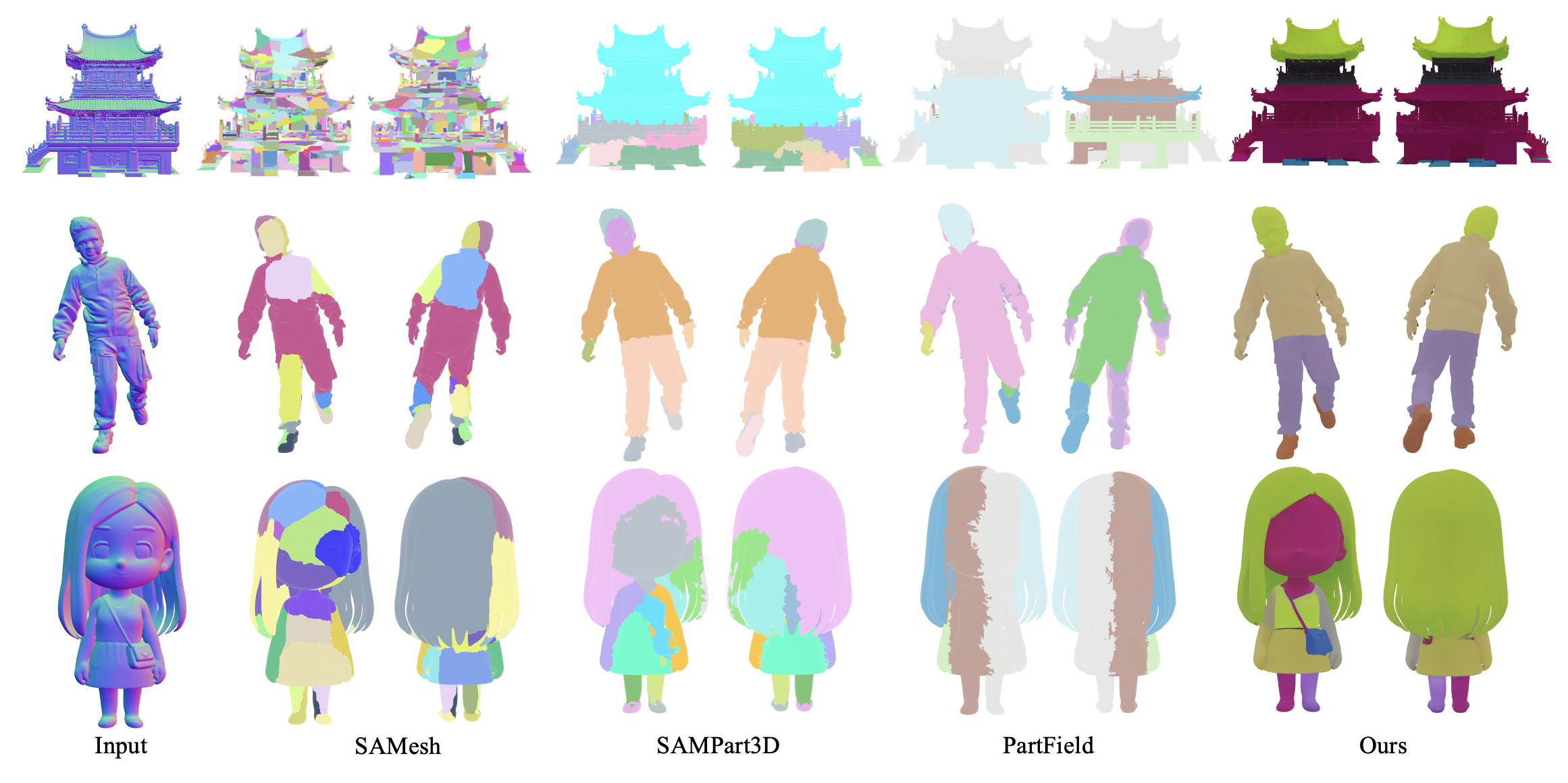}

  \caption{Qualitative comparison of part segmentation on high-polycount meshes. Our method (right) produces clean, sharp, and intuitively correct boundaries, while competing methods struggle with fragmented and unreasonable results.}
  \label{fig:seg_compare}
\end{figure*}

\paragraph{Multi-View  Texture Generation.} We also assess performance under multi-view conditions, comparing against leading commercial models in Figure~\ref{fig:commercial}. Our method consistently produces cleaner and more plausible textures in areas of significant occlusion. The generated texture demonstrates a more robust correspondence to the underlying geometry, fitting the mesh more accurately without the distortions or misalignments common in other methods.

\paragraph{Part Segmentation on Complex Meshes.} A core advantage of our unified framework is its extensibility to high-precision 3D perception. We evaluate 3D part segmentation, focusing on challenging, high-polycount meshes with complex topologies. We compare against prominent segmentation methods, including SAMesh \cite{tang2024segment}, SAMPart3D \cite{yang2024sampart3d}, and PartField \cite{partfield2025}. As shown in Figure \ref{fig:seg_compare}, competitor methods exhibit significant weaknesses on these complex assets. Their results are often fragmented, noisy, and lack clear boundaries, indicative of a failure to process the high-frequency geometric details. In contrast, our method produces exceptionally clean and sharp segmentation boundaries, which align strongly with human intuition, demonstrating our model's robust understanding of 3D structure, which is maintained by our attribute grid representation.

%

\subsection{Quantitative Analysis}

We provide quantitative metrics to substantiate our qualitative findings. We evaluate both front-view reconstruction fidelity and novel-view generative quality.

\begin{table}[htbp]
  \centering
  \caption{Quantitative comparison of texture generation and VAE reconstruction. Our method significantly outperforms competitors in all metrics.}
  \label{tab:quant_results}

  \setlength{\tabcolsep}{4pt}%

  \resizebox{\columnwidth}{!}{%

  \begin{tabular}{l ccc cc}
    \toprule

    \textbf{Method} & \multicolumn{3}{c}{\textbf{Front-View Reconstruction}} & \multicolumn{2}{c}{\textbf{Novel-View Generation}} \\

    \cmidrule(lr){2-4} \cmidrule(lr){5-6}

    & SSIM $\uparrow$ & PSNR $\uparrow$ & LPIPS $\downarrow$ & CLIP Score $\uparrow$  & CLIP FID $\downarrow$ \\
    \midrule

    Paint3D \cite{zeng2024paint3d} & 0.8903 & 21.5729 & 0.1051 & 0.8013  & 23.1599 \\
    TexGen \cite{texgen} & 0.8976 & 22.4177 & 0.1005 & 0.8206  & 22.1541 \\
    TRELLIS \cite{xiang2024structured} & 0.9150 & 25.5405 & 0.0856 & 0.8346 & 21.3961 \\
    \textbf{Ours} & \textbf{0.9421} & \textbf{30.0985} & \textbf{0.0627} & \textbf{0.8545}  & \textbf{19.8543} \\
    \bottomrule
  \end{tabular}%
  }
\end{table}
\vspace{-4mm}

\paragraph{Reconstruction and Generation Fidelity.}
As shown in Table \ref{tab:quant_results}, our method achieves significantly superior front-view reconstruction metrics (SSIM, PSNR, LPIPS) compared to all baselines. This highlights the powerful consistency enforced by our sparse latent conditioning, ensuring the generated texture faithfully adheres to the input view. Furthermore, our model also leads in novel-view generation metrics (CLIP Score, CLIP FID), demonstrating a state-of-the-art generative capability for rendering coherent and high-quality textures in unobserved regions.

\paragraph{Multi-View Conditioned Generation.}
We also evaluated our framework's synergy with multi-view methods by integrating it with MVAdapter (MVadapter+Ours). As shown in Table ~\ref{tab:method_comparison}, our combined approach outperformed all baselines on quantitative metrics for novel views. This confirms our framework can effectively enhance existing multi-view solutions, improving texture consistency and perceptual quality in unobserved viewpoints.
\begin{table}[htbp]
    \centering
    \caption{Quantitative Comparison over Multiview Texturing }
    \label{tab:method_comparison}
    \resizebox{\columnwidth}{!}{
    \begin{tabular}{lccc}
        \toprule
        Metric & Hunyuan2.0 & MVAdapter & \textbf{MVadapter+Ours} \\
        \midrule
        $\text{CLIP} \uparrow$ & 0.8124 & 0.8305& \textbf{0.8357} \\
        $\text{CLIP-FID} \downarrow$ & 23.1458 & 21.9120 & \textbf{21.7571} \\
        \bottomrule
    \end{tabular}
    }
\end{table}
\vspace{-10pt}

\begin{figure}[htbp]
  \centering
  \includegraphics[width=\columnwidth]{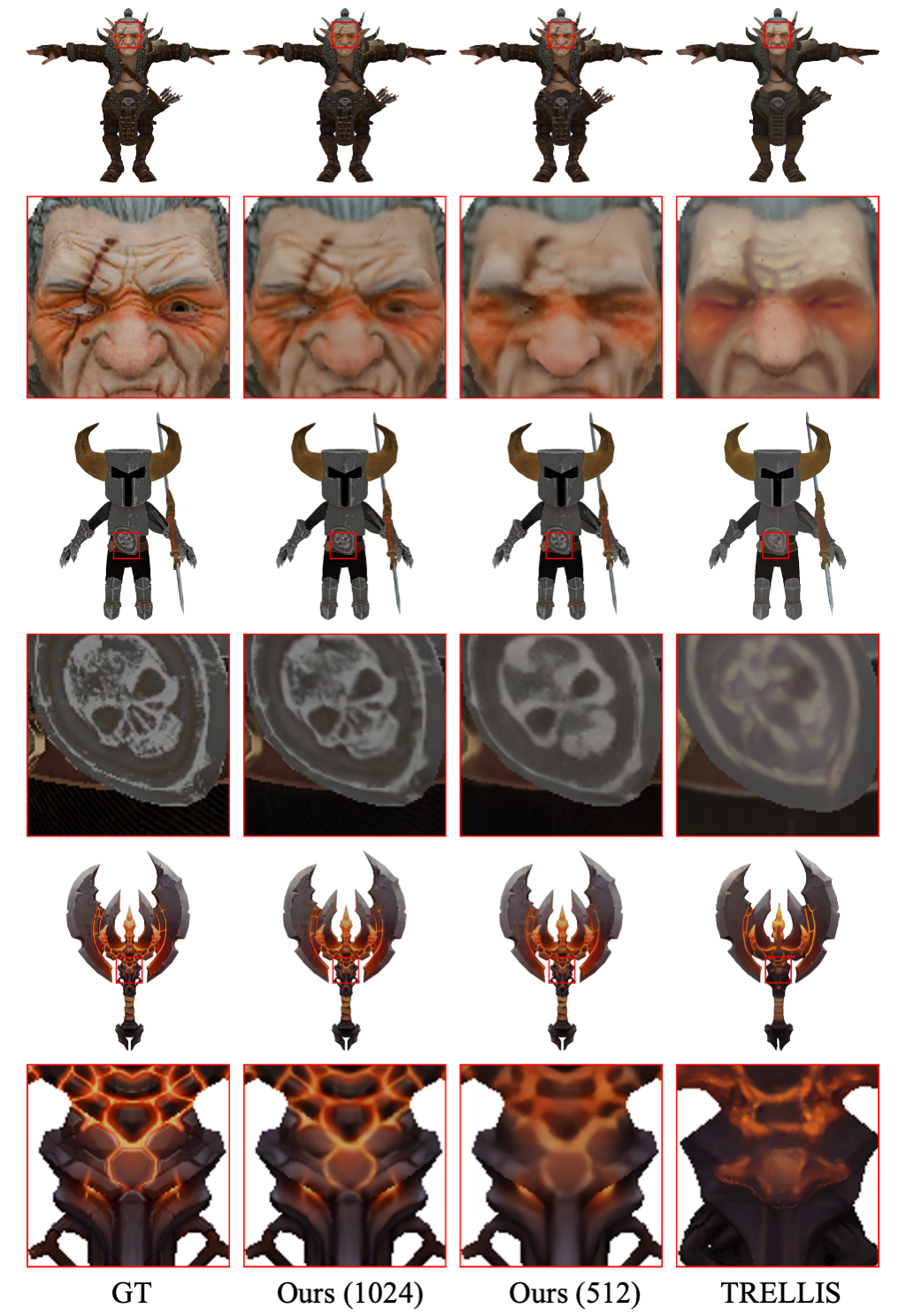}
    \caption{
  \textbf{Qualitative comparison of VAE reconstruction fidelity.}
  We compare our Attribute VAE against the VAE used in TRELLIS \cite{xiang2024structured}. 
  Our model (left) demonstrates a superior ability to reconstruct high-frequency texture details from the latent representation compared to TRELLIS (right).
}
  \label{fig:vae_comp}
\end{figure}

\paragraph{VAE Reconstruction Quality.}
We isolate the performance of our VAE by comparing its texture reconstruction fidelity against the VAE used in TRELLIS \cite{xiang2024structured} in Figure ~\ref{fig:vae_comp}. Our model produces results with sharper edges and better color fidelity compared to the baseline, even at a lower resolution. This efficient and high-fidelity latent representation is a cornerstone of our framework's overall performance.

\paragraph{Class-Agnostic Part Segmentation.}
We evaluate our model's class-agnostic part segmentation performance against SOTA methods: SAMesh~\cite{tang2024segment}, SAMPart3D~\cite{yang2024sampart3d}, and PartField~\cite{partfield2025}. We use two 100-mesh Objaverse subsets: \textbf{Objaverse (Random)} (randomly sampled) and \textbf{Objaverse (Complex)} (meshes with over 10,000 sharp edges, identified via DORA~\cite{Chen_2025_Dora}). Results are reported in mIoU ($\uparrow$) in Table~\ref{tab:segmentation_comparison}. As shown in Table~\ref{tab:segmentation_comparison}, our method achieves results comparable to SOTA (PartField) on the general benchmark. On the Objaverse (general), our grid-based query may introduce minor inconsistencies on the large, simple faces prevalent in this subset, which can span multiple voxel cells. However, on the Objaverse (Complex) subset, our approach demonstrates a significant advantage, outperforming all other methods by a large margin. This highlights our model's superior robustness in handling the intricate, high-frequency geometry that our high-resolution volumetric grid is designed to preserve.

\begin{table}[htbp]
    \centering
    \caption{Quantitative comparison for class-agnostic part segmentation (mIoU $\uparrow$). Best results are in \textbf{bold}.}
    \label{tab:segmentation_comparison}
    \resizebox{\columnwidth}{!}{
    \begin{tabular}{lcccc}
        \toprule
        Dataset & SAMesh & SAMPart3D & PartField & \textbf{Ours} \\
        \midrule
        Objaverse (Random) & 44.84 & 46.34 & \textbf{74.63} & 72.26 \\
        Objaverse (Complex) & 31.07 & 36.67 & 51.79 & \textbf{60.82} \\
        \bottomrule
    \end{tabular}
    }
\end{table}

\subsection{Ablation Study}

\begin{figure}[htbp]
  \centering
  \includegraphics[width=\columnwidth]{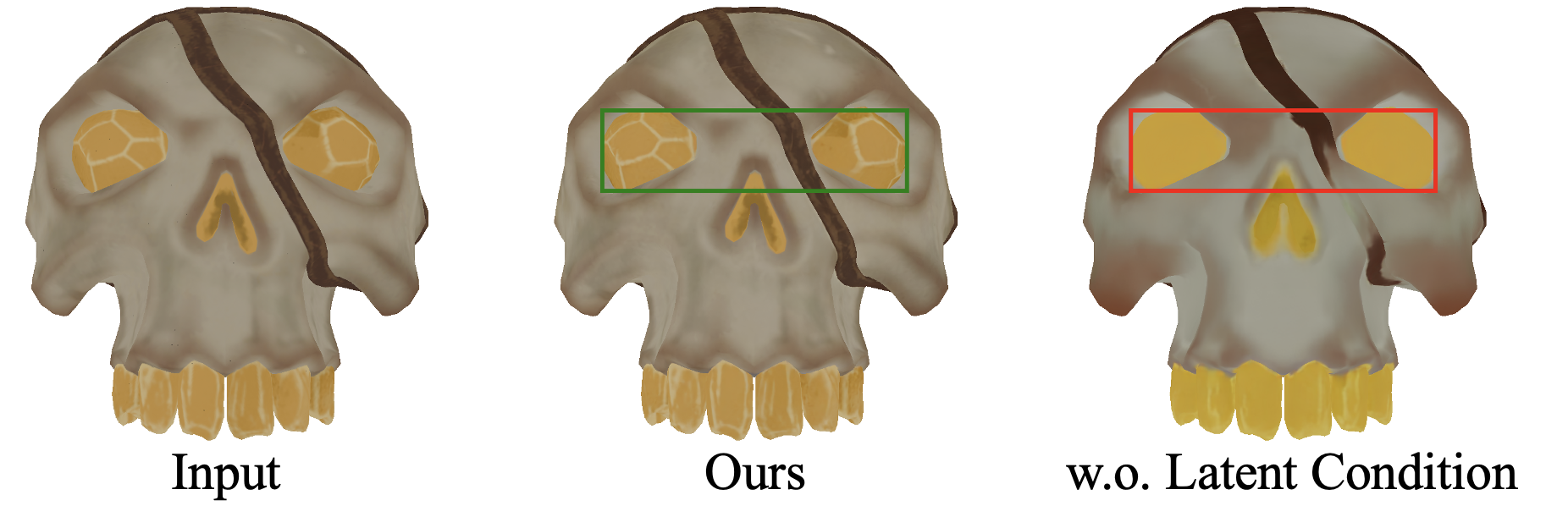}
  \caption{Ablation on Sparse Latent Condition. Our condition method yields more consistent results with the input image.}
  \label{fig:ablation_cond}
\end{figure}

\paragraph{Effect of Sparse Latent Condition.}
We train variants of our model to analyze the effect of our sparse latent condition derived from the input view. As shown in Table \ref{tab:ablation_dit}, removing this condition (`w/o Sparse Latent Condition`) results in a drastic degradation of front-view consistency, with SSIM and PSNR scores dropping significantly. Furthermore, the visualization in Figure ~\ref{fig:ablation_cond} clearly illustrates the impact of our conditioning strategy.. This result confirms that our sparse latent condition is critical for anchoring the generation process and ensuring high-fidelity alignment with the input. The table also compares single-view and multi-view conditional generation. 

\begin{table}[htbp]
  \centering
  \caption{Ablation study on the \textbf{DiT conditioning}. Removing the sparse latent condition significantly harms front-view consistency.}
  \label{tab:ablation_dit}
  \resizebox{\columnwidth}{!}{
  \begin{tabular}{l ccc}
    \toprule
    \textbf{Model Variant (DiT)} & SSIM $\uparrow$ & PSNR $\uparrow$ & LPIPS $\downarrow$ \\
    \midrule
    Ours (Single View) & \textbf{0.9421} & \textbf{30.0985} & \textbf{0.0627} \\
    w/o Sparse Latent Condition & 0.9052 & 24.3984 & 0.0942 \\
    \bottomrule
  \end{tabular}
  }
\end{table}
\vspace{-4mm}

\begin{figure}[htbp]
  \centering

  \includegraphics[width=\columnwidth]{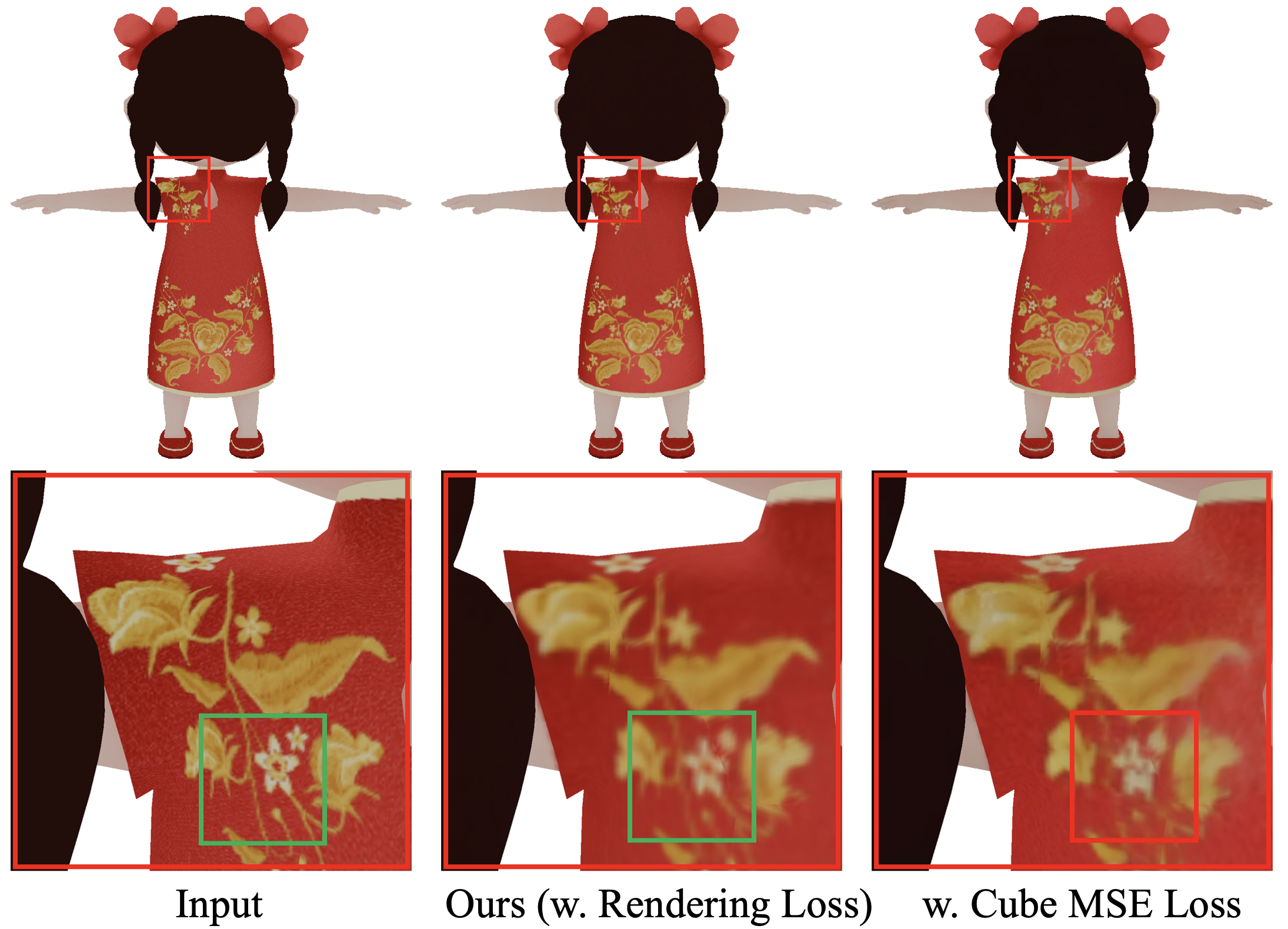}
  \caption{Ablation Study on Rendering Loss. Our VAE produces results with more details compared to cube MSE loss.}

  \label{fig:ablation_loss}
\end{figure}

\paragraph{Rendering-based vs. MSE Loss.}
We further investigate the impact of our training objective by comparing our rendering-based reconstruction loss against a more direct MSE (Mean Squared Error) loss computed directly on the sparse color grid (i.e., the "cube" loss). Although the MSE loss is computationally simpler, it treats all voxels equally, regardless of their visibility or contribution to the final rendered image. In contrast, our rendering-based loss operates in the image space. This approach is more aligned with human perception, forcing the model to optimize for the final visual projection. As shown in Figure \ref{fig:ablation_loss} and Table \ref{tab:ablation_loss}, the rendering-based loss yields results that are visually superior, with sharper details and fewer artifacts, and quantitatively better in terms of PSNR and SSIM metrics. 

\begin{table}[htbp]
  \centering
  \caption{Ablation study on the \textbf{VAE reconstruction loss}. The rendering-based loss achieves superior reconstruction fidelity.}
  \label{tab:ablation_loss}
  \resizebox{\columnwidth}{!}{
  \begin{tabular}{l ccc}
    \toprule
    \textbf{Model Variant (VAE)} & SSIM $\uparrow$ & PSNR $\uparrow$ & LPIPS $\downarrow$ \\
    \midrule
    Ours (w. Rendering Loss) & \textbf{0.9841} & \textbf{33.21} & \textbf{0.0246} \\
    w/ Cube MSE Loss & 0.9788 & 31.59 & 0.0421 \\
    \bottomrule
  \end{tabular}
  }
\end{table}
%
%

%% file: CVPR/section2/limit_and_conclusion.tex
\section{Limitation}

Despite the state-of-the-art results achieved by our framework in 3D generation and perception,  our framework's reliance on a latent attribute grid makes its performance sometimes sensitive to the quality and topology of the input mesh with mesh defects or severe self-intersections.
 
\section{Conclusion}
\label{sec:conclusion}

In this paper, we introduced a native 3D texturing framework that resolves multi-view inconsistency and UV fragmentation issues in previous 3D texturing methods. Our approach combines an Attribute Grid representation with a Sparse Latent Conditioning Diffusion Transformer, enabling native texture generation and perception. Experiments show that our framework reduces texture inconsistency and improves occluded regions, while delivering state-of-the-art performance.

\afterpage{
  \clearpage 
  \begin{figure*}[h!] 
    \centering 
    \includegraphics[height=0.85\textheight]{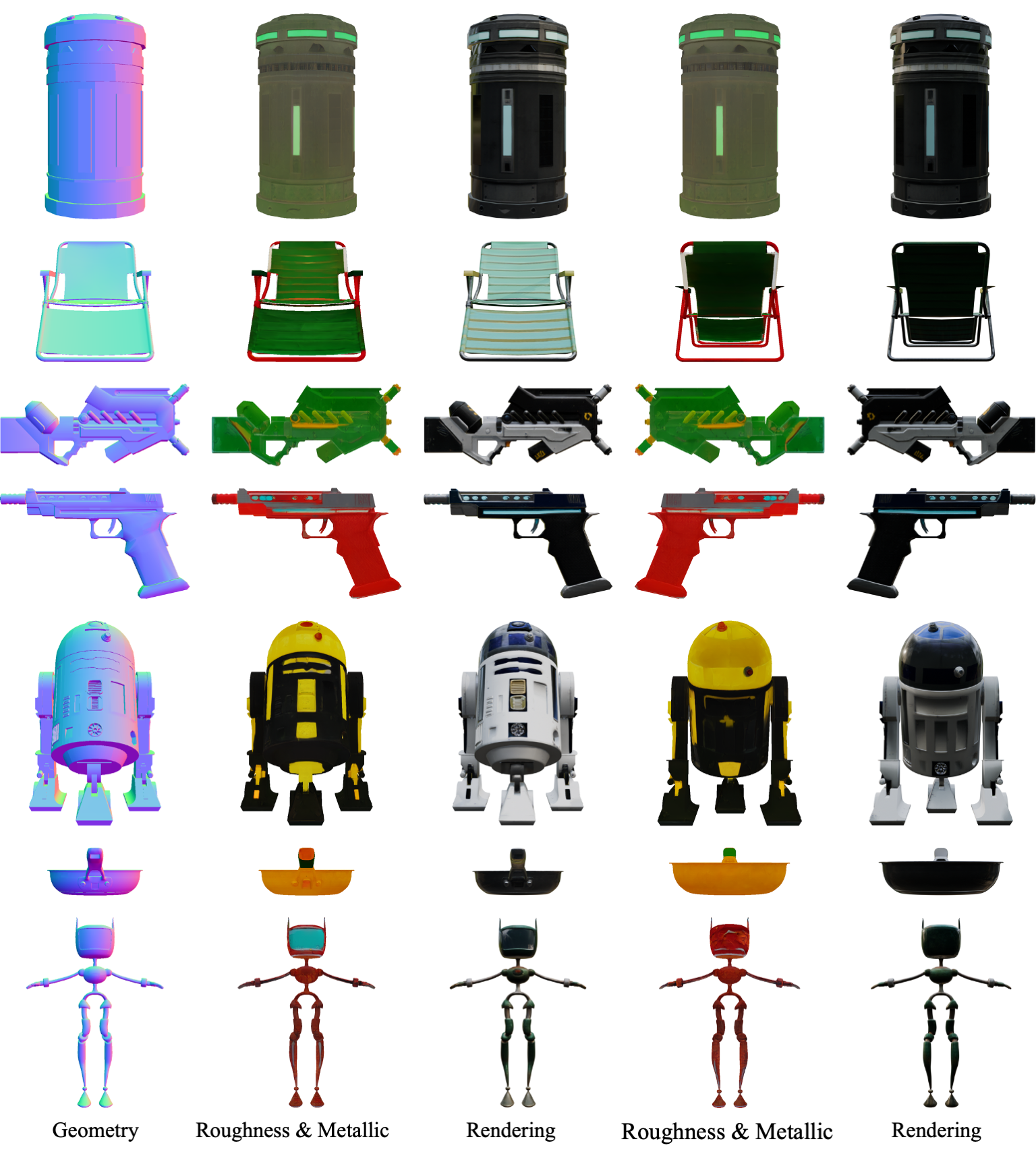}
    \centering
    \caption{Visualization for PBR Generation. Red Channel corresponds to Metallic and Green Channel corresponds to Roughness in the Roughness \& Metallic Column.}
    \label{fig:qual_pbr}
  \end{figure*}
  \clearpage 
}

\afterpage{
  \clearpage 
  \begin{figure*}[h!] 
    \centering 
    \includegraphics[height=0.96\textheight]{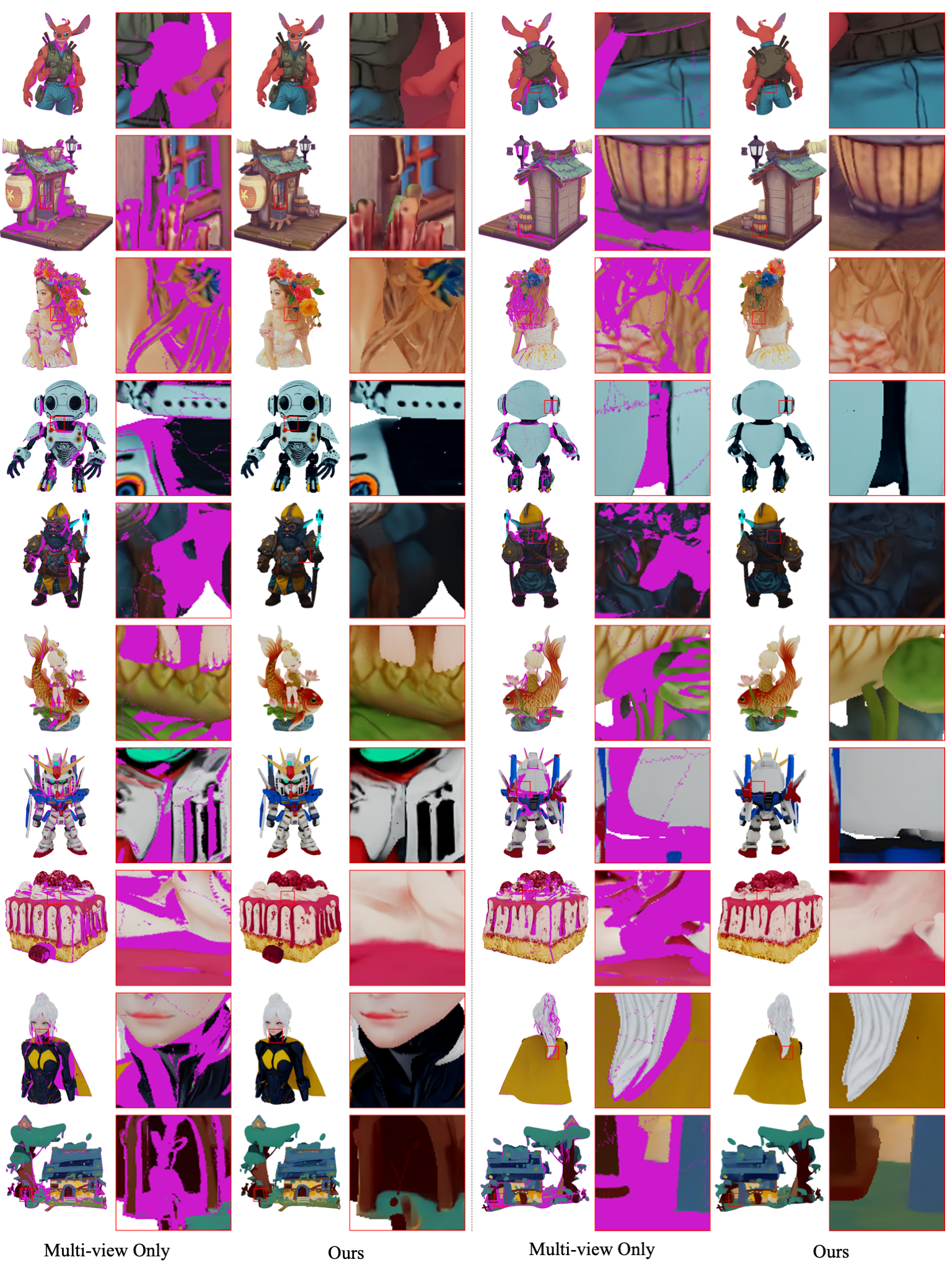}
    \caption{Visualization for More Examples of Multiview-Based Generation.}
    \label{fig:qual_mv}
  \end{figure*}
  \clearpage 
}

\afterpage{
  \clearpage 
  \begin{figure*}[h!] 
    \centering 
    \includegraphics[height=0.96\textheight]{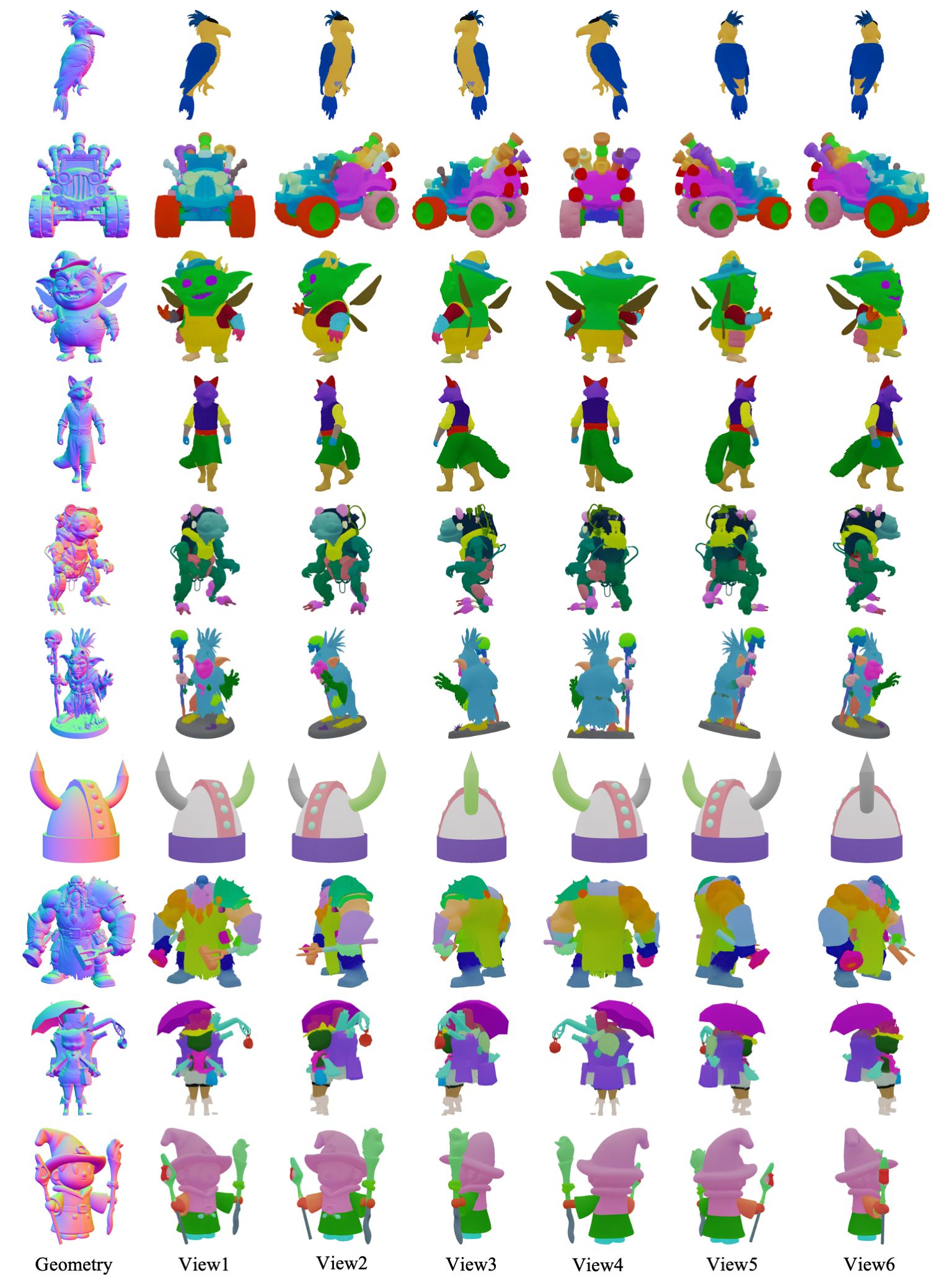}
    \caption{Visualization for More Examples of Our 3D Segmentation.}
    \label{fig:qual_seg}
  \end{figure*}
  \clearpage 
}